\documentclass[11pt]{article}

\usepackage{amsthm}
\newtheorem{theorem}{Theorem}[section]
\newtheorem{cor}[theorem]{Corollary}

\newtheorem{remark}{Remark}
\newtheorem{problem}{Problem}
\newtheorem{rem}[theorem]{Remark}

\newcommand{\WRP}{\par\qquad\(\hookrightarrow\)\enspace}

\usepackage{cite}
\usepackage{algorithm}
\usepackage[noend]{algpseudocode}
\usepackage{multirow}
\usepackage{color} 
\usepackage{amsfonts}
\usepackage{mysymbol}
\usepackage[dvipsnames]{xcolor}
\usepackage[hyphens]{url}
\usepackage{subcaption}

\usepackage{authblk}
\usepackage{setspace}
\usepackage{fullpage,etoolbox}
\usepackage{amsmath}
\usepackage{amssymb}
\allowdisplaybreaks
\newcommand{\vertiii}[1]{{\left\vert\kern-0.25ex\left\vert\kern-0.25ex\left\vert #1 
    \right\vert\kern-0.25ex\right\vert\kern-0.25ex\right\vert}}

\usepackage[breaklinks=true, colorlinks, bookmarks=true, citecolor=Black, urlcolor=Violet,linkcolor=Black]{hyperref}

\newcommand{\eps}{\varepsilon}

\usepackage{comment}
\usepackage[colorinlistoftodos,prependcaption,textwidth=1.5cm,textsize=tiny]{todonotes}

\title{Robust Motion Planning in the Presence of Estimation Uncertainty}
\author{Lars Lindemann}
\author{Matthew Cleaveland}
\author{Yiannis Kantaros}
\author{George J. Pappas\thanks{The authors are with the GRASP Laboratory, University of Pennsylvania, Philadelphia, PA 19104, USA. Email:{\tt\small \{larsl,mcleav, kantaros, pappasg\}@seas.upenn.edu} This material is based upon work supported by the Air Force Research Laboratory (AFRL) and the Defense Advanced Research Projects Agency (DARPA) under Contract No. FA8750-18-C-0090.}}
\affil{Department of Electrical and Systems Engineering, University of Pennsylvania}

\begin{document}

\maketitle

\begin{abstract}
Motion planning is a fundamental problem and focuses on finding control inputs that enable a robot to reach a goal region while safely avoiding obstacles.  However, in many situations, the state of the system may not be known but only estimated using, for instance, a Kalman filter.  This results in a novel motion planning problem where safety must be ensured in the presence of state estimation uncertainty. Previous approaches to this problem are either conservative or integrate state estimates optimistically which leads to non-robust solutions. Optimistic solutions require frequent replanning to not  endanger the safety of the system.  We propose a new formulation to this problem with the aim to be robust  to state estimation errors while not being overly conservative. In particular, we formulate a stochastic optimal control problem that contains robustified risk-aware safety constraints by incorporating robustness margins to account for state estimation errors. We propose a novel sampling-based approach that builds trees exploring the reachable space of Gaussian distributions that capture uncertainty both in state estimation and in future measurements. We provide robustness guarantees and show, both in theory and simulations, that the induced robustness margins constitute a trade-off between  conservatism and robustness for planning under estimation uncertainty that allows to control the frequency of replanning. 
\end{abstract}


\section{Introduction}

Motion planning is a fundamental problem that has received considerable research attention over the past years \cite{lavalle2006planning}. Typically, the motion planning problem aims to generate trajectories that reach a desired configuration starting from an initial configuration while avoiding unsafe states (e.g., obstacles). Several planning and control methods have been proposed to address this problem under the assumption that the unsafe state space is known; see e.g., \cite{karaman2010optimal,karaman2011sampling}. 

In this paper, we consider the problem of robust motion planning in the presence of state estimation uncertainty. In particular, we consider the case where the goal is to control a linear system using the Kalman filter to reach a desired final state while avoiding known unsafe states. To account for estimation uncertainty in current and predicted future system states due to noisy sensors and the Kalman filter,  we require the system's state to always respect risk-aware safety constraints.  First, we formulate this motion planning problem as a stochastic optimal control problem that generates control policies that, as a consequence of the Kalman filter, rely on future sensor measurements.
%
 Due to this dependence on future measurements, which are not available initially, we approximate this problem with an approximate stochastic optimal control problem that generates robust and uncertainty-aware control policies. Specifically, the generated policies are \textit{uncertainty-aware} in the sense that they take into account how estimation uncertainty may evolve under the execution of these controllers and the Kalman filter and \textit{robust} as they guarantee that risk-aware safety constraints are met even when the predicted state estimates deviate from the realized ones. In this paper, to solve this approximate stochastic optimal control problem, we build upon the $\text{RRT}^{*}$ algorithm \cite{karaman2011sampling} and we propose a new sampling-based approach that relies on searching the space of Gaussian distributions for the system's state. We provide correctness guarantees with respect to the approximate control problem and show how those guarantees relate to the original stochastic control problem. Our framework allows to integrate offline robust planning with online replanning. We argue, and show in simulations, that the robustness margins considered for offline planning constitute a fundamental trade-off between conservatism and robustness in the presence of state uncertainty that allows to control the frequency of replanning needed.

\textbf{Related Literature:} Motion planning under estimation uncertainty has been considered in \cite{kurniawati2012global,cai2019lets,sun2020stochastic} by using partially observable markov decision processes in belief space. As such approaches are often computationally intractable, sampling-based approximations have appeared in \cite{agha2014firm,vasile2016control,leahy2019control}. The authors in \cite{luders2010chance,luders2013robust} propose CC-RRT, an extension of \cite{karaman2011sampling}, that incorporates chance constraints to ensure probabilistic feasibility for linear systems subject to process noise. Extensions to account for uncertain dynamic obstacles are considered in \cite{kothari2013probabilistically,aoude2013probabilistically}. Common in these works is that the planning is not integrated with sensing and, therefore, these approaches are quite conservative as uncertainty may grow unbounded.
Sampling-based approaches that exhibit robustness to process noise have been proposed in \cite{luders2010bounds,summers2018distributionally}, but again without considering sensing information. The authors in \cite{safaoui2021risk} have used an unscented transformation to estimate state distributions for nonlinear systems within an RRT$^*$ framework by considering tightened risk constraints.  The authors in \cite{schuurmans2020learning,coulson2020distributionally} propose model predictive control frameworks that contain distributionally robust risk constraints that are based on ambiguity sets defined around an empirical state distribution from input-output data. 
%
On the other hand, sensor and sampling-based approaches that integrate Kalman or particle filters have been proposed in \cite{renganathan2020towards,berntorp2016particle,bry2011rapidly,van2011lqg}. These works have in common to treat future output measurements as random variables via its output measurement map, i.e., the map from states to observations. As a consequence, the estimated state becomes a random variable. This way, these methods lack robustness to uncertainty in output measurements. We instead impose robustness margins around a predicted trajectory under the Kalman filter and nominal output measurements, and replan if needed.


\textbf{Contribution:}  \textit{First}, we formulate the motion planning problem under state estimation uncertainty and risk constraints as a stochastic optimal control problem and propose a tractable approximation that is based on nominal predictions and robustified constraints. \textit{Second}, we propose a sampling-based algorithm towards solving the approximate problem and  show in what way our solution relates to the original problem. \textit{Third}, we show that our framework allows to integrate offline robust planning with online replanning. The robustness margins within the robustified constraints directly affect the frequency of replanning and hence constitute a fundamental trade-off between conservatism and robustness.

\section{Background}
\label{sec:backgound}

Let $\mathbb{R}$ and $\mathbb{N}$ be the set of real and natural numbers. Also let $\mathbb{N}_{\ge 0}$ be the set of non-negative natural numbers and $\mathbb{R}^n$ be the real $n$-dimensional vector space. Let $\mathcal{N}:\mathbb{R}^n\times \mathbb{R}^{n\times n}$ be an $n$-dimensional Gaussian distribution. 

\subsection{Random Variables and Risk Theory}

Consider the \emph{probability space} $(\Omega,\mathcal{F},P)$  where $\Omega$ is the sample space, $\mathcal{F}$ is a $\sigma$-algebra of $\Omega$, and $P:\mathcal{F}\to[0,1]$ is a probability measure. More intuitively, an element in  $\Omega$ is an \emph{outcome} of an experiment, while an element in $\mathcal{F}$ is an \emph{event} that consists of one or more outcomes whose probabilities can be measured by the probability measure $P$. 

\textbf{Random Variables.} Let $X$ denote a real-valued \emph{random vector}, i.e., a measurable function $X:\Omega\to\mathbb{R}^n$. We refer to $X(\omega)$ as a realization of  $X$ where $\omega\in\Omega$. As $X$ is  measurable, a probability space can be defined for $X$ and probabilities can be assigned to events for values of $X$. 


\textbf{Risk Theory.} Let $\mathfrak{F}(\Omega,\mathbb{R})$ denote the set of measurable functions mapping from the domain $\Omega$ into the domain $\mathbb{R}$. A \emph{risk measure} is a function $R:\mathfrak{F}(\Omega,\mathbb{R})\to \mathbb{R}$ that maps from the set of real-valued random variables to the real numbers.  Risk measures  allow for a risk assessment when the input of $R$ is seen as a cost random variable. Commonly used risk measures are the expected value, the variance, or the conditional value-at-risk \cite{rockafellar2000optimization}. In Appendix \ref{app:risk}, we summarize existing risk measures and their properties.

\subsection{Stochastic Control System}
Consider the discrete-time stochastic control system
\begin{subequations}\label{eq:rdynamics}
\begin{align}
X(t+1)&=A X(t) + Bu(t)+ W(t), \;X(0):=X_0\\
Y(t)&=CX(t)+ V(t)\label{eq:rdynamics_y}
\end{align}
\end{subequations}
where $X(t)\in\mathbb{R}^{n}$, $u(t)\in\ccalU \subseteq\mathbb{R}^{m}$, $Y(t)\in\mathbb{R}^{p}$ denote the state of the system, the control input, and the measurement at time $t$. The set $\ccalU$ denotes the set of admissible control inputs. Also, $W(t)\in\mathbb{R}^n$ and $V(t)\in\mathbb{R}^p$ denote the state disturbance and the measurement noise at time $t$ and they are assumed to follow a Gaussian distribution, i.e., $ W(t)\sim\ccalN(\mu_W,\Sigma_W)$ and $V(t)\sim\ccalN(\mu_V,\Sigma_V)$,
with known mean vectors $\mu_W$ and $\mu_V$ and known covariance matrices $\Sigma_W$ and $\Sigma_V$. The initial condition $X_0\in\mathbb{R}^n$  also follows a Gaussian distribution, i.e,  $ X_0\sim\ccalN(\mu(0),\Sigma(0))$.
We assume that $X(0)$, $W(t)$, and $V(t)$ are mutually independent. Note that we have here dropped the underlying sample space $\Omega$ of $W(t)$, $V(t)$, and $X(0)$ for convenience.

\subsection{Kalman Filter for State Estimation}
Note  that the state $X(t)$ and the output measurements $Y(t)$ also become Gaussian random variables  since the system in \eqref{eq:rdynamics} is linear in $W(t)$, $V(t)$, and $X(t)$. The state $X(t)$ defines a stochastic process with a mean $\mu(t)$ and a covariance matrix $\Sigma(t)$ that can recursively be calculated as stated in Appendix \ref{app:state_est}. The estimates $\mu(t)$ and $\Sigma(t)$  are conservative and do not incorporate available output measurements, i.e., the realizations $y(t)$ of $Y(t)$. Let us denote the realized output measurements up until time $t$ as $Y_{t}:=\begin{bmatrix} y(0)^T &\hdots & y(t)^T\end{bmatrix}^T.$
We can now refine the above estimates to obtain optimal estimates based on $Y_{t}$ by means of a Kalman filter. For $s\le t$, let us for brevity define the random variable $X(t|s):=X(t)|Y_{s}$
as the random variable $X(t)$ conditioned on knowledge of the realized output measurements $Y_{s}$ with the conditional mean $\mu(t|s)$ and the conditional covariance matrix $\Sigma(t|s)$. We can calculate these quantities recursively as
\begin{subequations}\label{eq:kalman_eq}
\begin{align}
        \mu(t+1|t+1)&=F_\mu(\mu(t|t),u(t),y(t+1))\\
    \Sigma(t+1|t+1)&=F_\Sigma(\Sigma(t|t))\label{eq:kalman_eq_}
\end{align}
\end{subequations}
where the functions $F_\mu:\mathbb{R}^n\times\mathbb{R}^m\times\mathbb{R}^p\to\mathbb{R}^n$ and $F_\Sigma:\mathbb{R}^{n\times n}\to\mathbb{R}^{n\times n}$  are defined in Appendix \ref{app:state_est}.

\begin{rem}\label{rem:2}
 It holds that $\mu(t)=\mu(t|t)$ for $t\in\mathbb{N}_{\ge 0}$ if and only if the innovation term within the Kalman filter is zero for all $t\in \mathbb{N}_{\ge 0}$, i.e., if $y(t)=C\mu(t|t-1)$ for all $t\in \mathbb{N}_{\ge 0}$. 
 However, if the innovation term is not always zero, i.e., if $y(t)\neq C\mu(t|t-1)$ for some $t\in\mathbb{N}_{\ge 0}$, then $\mu(t)\neq\mu(t|t)$  for at least one time $t\in\mathbb{N}_{\ge 0}$. 
\end{rem}

\section{Problem Formulation}
\label{problem_formulation}
\subsection{Risk-Aware Stochastic Optimal Control Problem}
Consider a stochastic system of the form \eqref{eq:rdynamics} operating in a compact environment occupied by $M$ regions of interest $O_i\in \mathbb{R}^n$ for $i\in\ccalM:=\{1,\dots,M\}$. The goal is to safely navigate the system to the goal region $O_1$ by means of the control input $u(t)$. Additionally, the system has satisfy spatial requirements, i.e., to avoid and/or be close to regions $O_2, \hdots, O_M$. For instance, consider a robot that always has to avoid obstacles and  be within communication range of a static wifi spots. Consider therefore $J$ measurable functions $d_j:\mathbb{R}^{nM}\to\mathbb{R}$ for $j\in\ccalJ:=\{1,\dots,J\}$ 
that will encode such constraints.  We aim to evaluate the function $d_j$ based on the conditional state estimate $X(t|t)$. Since $X(t|t)$ is a random variable, the functions $d_j(X(t|t),O_2,\hdots,O_M)$ also become random variables.  The problem that this paper addresses can be captured by the following \textit{stochastic} optimal control problem:
\begin{subequations}
\label{eq:Prob1}
\begin{align}
 \min_{\substack{H, U_H}} &\sum_{t=0}^{H}  c(\mu(t|t)) \label{obj2}\\
 \text{s.t. } & \mu(t+1|t+1)=F_\mu(\mu(t|t),u(t),y(t+1)), \forall t\in \mathcal{T}_H\label{constr2}\\
&  \Sigma(t+1|t+1)=F_\Sigma(\Sigma(t|t)), \forall t\in \mathcal{T}_H\label{constr2_c}\\
& R(-d_j(X(t|t),O_2,\hdots,O_M))\leq\gamma_j, \forall j\in\ccalJ, \forall t\in \mathcal{T}_H \label{constr2b}\\
&   R(\|X(H|H)-O_1\|- \kappa)\le \gamma,  \label{constr1b} 
\end{align}
\end{subequations}
where $U_t:=\begin{bmatrix} u(0)^T &\hdots & u(t)^T\end{bmatrix}^T$, $\mathcal{T}_H=\{0,\hdots,H\}$, $c:\mathbb{R}^n\to \mathbb{R}_{\ge 0}$ is a  cost function, $\kappa>0$ indicates the accuracy for reaching $O_1$, and $\gamma$ and $\gamma_j$ are given risk thresholds.

\subsection{The Approximate Stochastic Optimal Control Problem}
The challenge in solving the optimization problem \eqref{eq:Prob1} lies in the dependence of the cost function \eqref{obj2} and the constraints \eqref{constr2}, \eqref{constr2b}, and \eqref{constr1b}  
on the realized output measurements  $y(t)$. Therefore, \eqref{eq:Prob1} can not be solved a priori. 

Unlike \cite{renganathan2020towards,berntorp2016particle,bry2011rapidly,van2011lqg} where $y(t)$ is treated as a random variable, consequently making $\mu(t|t)$ itself a random variable, we  approximate the optimization problem \eqref{eq:Prob1} with an approximate stochastic optimal control problem by using $\mu(t|0)$ instead of $\mu(t|t)$. Recall that $\mu(t|0)$ is computed using only the prediction step of the Kalman filter so that $\mu(t|0)$ is equivalent to the unconditional mean $\mu(t)$. To account for a potential mismatch between  $\mu(t|0)$ and the realization of $\mu(t|t)$, we introduce additional \textit{robustness} margins to alleviate the lack of knowledge of $Y_t$ during planning.  

In particular, we account for all realizations $y(t)$ that are such that $X(t|t)$ is '$\epsilon$-close' to the random variable $\hat{X}(t|t)\sim\mathcal{N}(\mu(t|0),\Sigma(t|t))$ that we use for planning.\footnote{Closeness here is in terms of the 2nd Wasserstein distance between $X(t|t)$ and $\hat{X}(t|t)$. Note in particular that the 2nd Wasserstein distance between $X(t|t)$ and $\hat{X}(t|t)$ is equivalent to $\|\mu(t|0)-\mu(t|t)\|^2$ as both $X(t|t)$ and $\hat{X}(t|t)$ are Gaussian with the same covariance matrix.} To formalize the '$\epsilon$-closeness' notion, define the set of distributions
\begin{align*}
\mathcal{B}_{\epsilon}\big(\hat{X}(t|t)\big):=\big\{\mathcal{N}\big(\mu,\Sigma(t|t)\big)|\exists {\mu}\in\mathbb{R}^n, \|{\mu}-\mu(t|0)\|^2\le \epsilon \big\}
\end{align*}
which contains all distributions with a covariance of $\Sigma(t|t)$ in an Euclidean ball of size $\epsilon$ around $\mu(t|0)$ where $\epsilon$ is a design robustness parameter. As it will be shown in Section \ref{sec:guarantees}, the size of $\epsilon$ will determine the probability by which the constraints in \eqref{eq:Prob1} are satisfied. In particular, we let $\epsilon:\mathbb{R}^{n \times n}\times \mathbb{N}_{\ge 0}\to \mathbb{R}_{\ge 0}$ be a function that depends on $A$ and $t$, i.e., $\epsilon(A,t)$. We will drop the dependence on $A$ and $t$  when it is clear from the context for ease of notation. Naturally, the discrepancy between $X(t|t)$ and $\hat{X}(t|t)$ increases with time so that a larger $\epsilon$ may be desired for larger $t$. 



We approximate the stochastic optimal control problem \eqref{eq:Prob1} with the  stochastic optimal control problem
\begin{subequations}
\label{eq:Prob2}
\begin{align}
& \min_{\substack{H, U_H}} \sum_{t=0}^{H}  c(\mu(t|0)) \label{obj2_}\\
\text{s.t. }&  \mu(t+1|0)=A\mu(t|0)+Bu(t)+\mu_W,\; \forall t\in \mathcal{T}_H \label{constr2_}\\
&  \Sigma(t+1|t+1)=F_\Sigma(\Sigma(t|t)), \;\forall t\in \mathcal{T}_H\label{constr2_c_}\\
\begin{split}
&   \sup_{{X} \in\mathcal{B}_{\epsilon(A,t)}(\hat{X}(t|t))}R(-d_j({X},O_2,\hdots,O_M))\leq\gamma_j,\;\forall j\in\ccalJ, \forall t\in \mathcal{T}_H \label{constr2b_}
\end{split}\\
&  \sup_{{X} \in\mathcal{B}_{\epsilon(A,H)}(\hat{X}(H|H))}R(\|{X}-O_1\|- \kappa)\le \gamma,  \label{constr1b_} 
\end{align}
\end{subequations}
where $\hat{X}(t|t)\sim\mathcal{N}(\mu(t|0),\Sigma(t|t))$. 
\begin{problem}\label{prob:1}
Given the system in \eqref{eq:rdynamics} and the regions of interest $O_i$, determine a terminal horizon $H$ and a sequence of control inputs $U_H$ that solves \eqref{eq:Prob2}.
\end{problem}

\begin{rem}[Approximation Gap]
Note that a solution to \eqref{eq:Prob2} may not constitute a feasible solution to \eqref{eq:Prob1}. The reason is that \eqref{eq:Prob1} relies on knowledge of measurements that will be taken in the future which is not the case in \eqref{eq:Prob2}. Although, this is partially accounted for by introducing the robustness parameter $\epsilon$, the feasibility gap may still exist depending on the values of $\epsilon$. In Section \ref{sec:guarantees}, we show that the probability that a solution to \eqref{eq:Prob2} satisfies the constraints of \eqref{eq:Prob1} depends on $\epsilon$. Moreover, to further account for this feasibility gap, in Section \ref{sec:online}, we propose a re-planning framework that is triggered during the execution time when the robustness requirement ${X}(t|t) \in\mathcal{B}_{\epsilon(A,t)}(\hat{X}(t|t))$ is not met.
\end{rem}

\begin{rem}[Robustness]
Note that in \eqref{eq:Prob2}, although the main purpose of the robustness parameter $\epsilon$ is to account for uncertainty in the measurements that will be collected online, due to its generality it also provides robustness to other modeling uncertainties e.g., inaccurate models of the system, the process, or the measurement noise.
\end{rem}

\begin{rem}[Sensor Model]
The sensor model described in \eqref{eq:rdynamics} can model e.g., a GPS-sensor, used for estimating the system's state, e.g., the position of a robot. By definition of this sensor model, it is linear and it does not interact with the environment. Nevertheless, more complex nonlinear sensor models that allow for environmental interaction can be considered (e.g., a range sensor). To account for such sensor models, an Extended Kalman filter (EKF) can be used. In this case, the constraints \eqref{constr2_c_} require linearization of the sensor model with respect to the hidden state; see e.g., \cite{atanasov2014information}. 
\end{rem}

\section{Robust Rapidly Exploring Random Tree (R-RRT$^*$)}
\label{sec:rrt}
We build upon the RRT$^*$ algorithm \cite{karaman2011sampling} and present a new robust sampling-based approach to solve \eqref{eq:Prob2}. In particular, in Sections \ref{sec:tree} and \ref{sec:increment} we propose an offline planning algorithm that relies on building trees incrementally that simultaneously explore the reachable space of Gaussian distributions modeling the system's state. In Section \ref{sec:online}, we propose a replanning algorithm for cases when $X(t|t)\not\in \mathcal{B}_{\epsilon}(\hat{X}(t|t))$ during the online execution of the algorithm.

\subsection{Tree Definition}
\label{sec:tree}

In what follows, we denote the constructed directed tree by $G:=(\mathcal{V},\mathcal{E})$, where $\ccalV\subseteq \mathbb{R}^n \times \mathbb{R}^{n\times n}\times \mathbb{N}_{\ge 0}$ is the set of nodes and $\mathcal{E}\subseteq \mathcal{V}\times\mathcal{V}\times \mathcal{U}$ denotes the set of edges. The set $\mathcal{V}$ collects nodes $v$ consisting of a nominal mean and a nominal covariance matrix that are denoted by $v.{\mu}$ and $v.{\Sigma}$, respectively, along with a time stamp $v.t$. We remark that each node will accept state distributions $X$ that are such that $X\in \mathcal{B}_{\epsilon}\big(\mathcal{N}(v.{\mu},v.{\Sigma})\big)$. The set $\mathcal{E}$ collects edges from a node $v$ to another node $v'$ along with a control input $u$, if the state distribution in $v'$ is reachable from $v$ at time $v.t$ and by means of $u$  according to the dynamics in \eqref{eq:kalman_eq}. The cost of reaching a node $v'$ with parent node $v$ is
\begin{equation}\label{eq:costUpd}
\text{Cost}(v')= \text{Cost}(v) +  c(v'.\mu).
\end{equation}
Observe that by applying \eqref{eq:costUpd} recursively, we get that $\text{Cost}(v')$ is the objective function in \eqref{eq:Prob2}.
The tree is rooted at a node $v_0$ capturing the initial system state, i.e., $v_0.\mu=\mu(0|0)$, $v_0.\Sigma=\Sigma(0|0)$, $v_0.t=0$, and $\text{Cost}(v_0)=c(v_0.\mu)$. Construction of the tree occurs in an incremental fashion, as detailed in Algorithm \ref{alg:1}, where within a single iteration (i) a new node is sampled, (ii) the tree is extended towards this new sample, if possible, and (iii) a rewiring operation follows aiming to decrease the cost of existing tree nodes by leveraging the newly added node. 
After taking $N\geq 0$ samples [line \ref{alg1:forIter}, Alg. \ref{alg:1}], where $N$ is user-specified, Algorithm \ref{alg:1} terminates and returns a solution to  \eqref{eq:Prob2}, if it has been found,  i.e., a terminal horizon $H$ and a sequence of control inputs $U_H$. 

To extract such a solution, we first need  to define the set $\mathcal{V}_G\subseteq \mathcal{V}$ that collects all nodes $v\in \mathcal{V}$ of the tree that satisfy the terminal constraint \eqref{constr1b_}. 
Then, among all nodes in $\mathcal{V}_G$, we select the node $v\in \mathcal{V}_G$ with the smallest cost $C(v)$.
Then, the terminal horizon is $H:=v.t$ and the control inputs $U_H$ are recovered by computing the path in $G$ that connects the selected node $v\in \mathcal{V}_G$ to the root $v_0$. Note that satisfaction of the remaining constraints in \eqref{eq:Prob2} is guaranteed by construction of the tree $G$. A detailed description follows next. 

\subsection{Incremental Construction of Tree}
\label{sec:increment}
\textsc{Sample}: At every iteration $i$ of Algorithm \ref{alg:1}, we first  generate a sample from the space of Gaussian distributions modelling the system's uncertain state. To sample  from the set of Gaussian distributions, we define the function $\textsc{Sample}:\mathbb{N}_{\ge 0}\times \Omega \to \mathbb{R}^n\times S_{\ge 0}$ that generates independent and identically distributed  samples, denoted by $s_{\text{rand}}$, of means $\mu$ and covariances $\Sigma$ where $S_{\ge 0}\subseteq \mathbb{R}^{n\times n}$ is the set of positive semidefinite matrices [line \ref{alg1:sample}, Alg. \ref{alg:1}]. For simplicity, we use $\textsc{Sample}(i)$ instead of $\textsc{Sample}(i,\omega)$ and again omit the underlying sample space $\Omega$. 
\begin{rem}[Sampling]
 RRT$^*$ algorithms typically draw samples directly from the obstacle-free space. This is not possible here due to the stochastic setup so that nodes consist of means and a covariances. These nodes need to be dynamically feasible according to \eqref{eq:kalman_eq} with respect to the parent node  while satisfying the constraint \eqref{constr2b_}. We sample both means and covariances despite having no control over the covariance update equation \eqref{eq:kalman_eq_}. This follows as we try to connect new nodes to existing nodes $v\in\mathcal{V}$ that may be different time hops away from the root node $v_0$.
\end{rem}

\textsc{Nearest}: Next, given the sample $s_{\text{rand}}\in\mathbb{R}^n\times S_{\ge 0}$, among all nodes in the current tree structure, we pick the closest one to $s_{\text{rand}}$, denoted by $v_\text{nearest}$ [line \ref{alg1:nearest}, Alg. \ref{alg:1}]. To this end, we define the following function:
\begin{align*}
   \textsc{Nearest}(\mathcal{V},\mathcal{E},s_{\text{rand}}):=\argmin_{v\in \mathcal{V}}\; d(\mathcal{N}(s_{\text{rand}}),\mathcal{N}(v.\mu,v.\Sigma))
\end{align*}
where $d(\cdot)$ is the 2nd Wasserstein distance.\footnote{Particularly, we define $d(\mathcal{N}(\mu_1,\Sigma_1),\mathcal{N}(\mu_2,\Sigma_2)):=\|\mu_1-\mu_2\|^2+\text{trace}(\Sigma_1+\Sigma_2-2(\Sigma_2^{\frac{1}{2}}\Sigma_1\Sigma_2^{\frac{1}{2}})^{\frac{1}{2}})$.}

\textsc{Steer}: 
In order to steer the tree from a node $v$ towards $s$ [lines \ref{alg1:steer}, \ref{alg1:steer_}, and \ref{alg1:steer__}, Alg. \ref{alg:1}], we use the function $\textsc{Steer}(v,s,\eta):=(v_\text{new},u_\text{new}),$
where $\eta\ge 0$ is a positive constant, and returns the closest to $s$ dynamically feasible node $v_\text{new}:=({\mu}_\text{new},{\Sigma}_\text{new},v.t+1)$ that is still $\eta$-close to the node $v$. In particular, the mean ${\mu}_\text{new}$ and the covariance ${\Sigma}_\text{new}$ are required to be dynamically feasible according to the dynamics in \eqref{eq:kalman_eq}. The steer function aims to solve the optimization problem
\begin{subequations}
\begin{align}
    ({\mu}_\text{new},{\Sigma}_\text{new},u_\text{new})&:=\argmin_{({\mu},\Sigma,u)\in\mathbb{R}^n\times S_{\ge 0}\times\mathbb{R}^m} d(\mathcal{N}(s),\mathcal{N}(\mu,\Sigma))\nonumber\\
    \text{s.t. }& d(\mathcal{N}(v.\mu,v.\Sigma),\mathcal{N}(\mu,\Sigma))\le \eta\nonumber\\
    &{\mu}=Av.{\mu}+Bu+\mu_W \label{eq:steer1}\\
    &{\Sigma}=F_\Sigma(v.\Sigma)\label{eq:steer2}\\
    & u\in \mathcal{U}\label{eq:steer3}.
\end{align}
\end{subequations}
The above optimization program can be rewritten as a convex optimization program if $\mathcal{U}$ is a convex set. First, note that $\Sigma=F_\Sigma(v.\Sigma)$ is fixed according to constraint \eqref{eq:steer2}. For this $\Sigma$, the expressions involving $d(\cdot)$ then result in expressions that are convex in $\mu$ due to the use of the 2nd Wasserstein distance. Finally, \eqref{eq:steer1} and \eqref{eq:steer3} are convex in $u$ if $\mathcal{U}$ is convex.

\textsc{CheckConstrRob}: We next check if the new node $v_\text{new}$ is feasible i.e.,  if the constraint \eqref{constr2b_} holds for all distributions accepted by $v_\text{new}$ [line \ref{alg1:CheckConstrRob}, Alg. \ref{alg:1}]., i.e., if 
\begin{align}\label{eq:rob_check}
   \sup_{X\in \mathcal{B}_{\epsilon(A,v_{\text{new}}.t)}(\hat{X})} R(-d_j(X,O_2,\hdots,O_M))\leq\gamma_j, \forall j\in\ccalJ.
\end{align}
where $\hat{X}\sim \mathcal{N}(v_{\text{new}}.\mu,v_{\text{new}}.\Sigma)$. Checking \eqref{eq:rob_check} analytically may in general be hard. Sampling-based solutions as in \cite{rockafellar2000optimization} for the the conditional value-at-risk (CVaR) or \cite{lindemann2021stl} for logic specifications can be used, while there even exist sampling-based reformulations, again for the CVaR in  \cite[Sec. IV]{coulson2020distributionally}.

\textsc{Near}: If the transition from $v_{\text{nearest}}$ to $v_{\text{new}}$ is feasible, we check if there is any other candidate parent node for $v_{\text{new}}$ that can incur to $v_{\text{new}}$ a lower cost than the one when $v_{\text{nearest}}$ is the parent node. The candidate parent nodes are selected from the following set [line \ref{alg1:Vnear}, Alg. \ref{alg:1}]
\begin{align}\label{eq:near}
    &\textsc{Near}(\mathcal{V},\mathcal{E},s_{\text{new}},\eta):=\{v\in V|\|\mathcal{N}(s_{\text{new}})-\mathcal{N}(v.\mu,v.\Sigma)\|\le r(\mathcal{V})\}.
\end{align}
where $r(\ccalV)$ is, for $\gamma>0$, defined as
\begin{align*}
r(\mathcal{V}):=\min\Big\{\gamma\Big(\frac{\log\left|\mathcal{V}\right|}{\left|\mathcal{V}\right|}\Big)^{\frac{1}{n}}, \eta\Big\}.
\end{align*}
The set in \eqref{eq:near} collects all nodes $v\in \mathcal{V}$ that are within at most a distance of $r$ from $s_{\text{new}}$ in terms of the norm $\|\cdot\|$. Among all candidate parents, we pick the one that incurs the minimum cost while ensuring that transition from it towards $v_\text{new}$ is feasible; the selected parent node is denoted by $v_\text{new}$ [lines \ref{alg1:Vnear}-\ref{alg1:end1}, Alg. \ref{alg:1}]. Next, the sets of nodes and edges of the tree are accordingly updated [lines \ref{alg1:new_node}-\ref{alg1:new_edge}, Alg. \ref{alg:1}].


\textsc{GoalReached}: Once a new node $v_{\text{new}}$ is added to the tree, we check if its respective Gaussian distribution  satisfies the terminal constraint \eqref{constr1b_}. This is accomplished by the function $\textsc{GoalReached}$ [line \eqref{alg1:GoalReached}, Alg. \ref{alg:1}]. This function resembles the function \textsc{CheckConstrRob}, but it focuses on the terminal constraint \eqref{constr1b_} instead of \eqref{constr2b_}. Specifically, for a given node $v_{\text{new}}\in V$, the function $\textsc{GoalReached}(v_\text{new})$ checks robust satisfaction of the goal constraint, i.e., whether or not the following condition is met:
\begin{align}\label{eq:rob_check_}
   \sup_{X\in \mathcal{B}_{\epsilon(A,v_{\text{new}}.t)}(\hat{X})} R(\|X-O_1\|-\kappa)\leq\gamma.
\end{align}
where $\hat{X}\sim \mathcal{N}(v_{\text{new}}.\mu,v_{\text{new}}.\Sigma)$. If this condition is met, the set $\ccalV_G$ is updated accordingly [line \eqref{alg1:updateGoalSet}, Alg. \ref{alg:1}].

Finally, given a new node $v_{\text{new}}$, we check if the cost of the nodes collected in the set \eqref{eq:near} can decrease by rewiring them to $v_{\text{new}}$, as long as such transitions respect \eqref{constr2b_} (or equivalently \eqref{eq:rob_check}) [lines \ref{alg1:rew_start}-\ref{alg1:rew_end}, Alg. \ref{alg:1}]. 

\textsc{CheckConstrRew}:  Note that the constraint in \eqref{eq:rob_check} is checked for the set of distributions $\mathcal{B}_{\epsilon(A,v_\text{new}.t)}(\hat{X})$ when a node $v_\text{new}\in \mathcal{V}$ is added to the tree. If now a node $v_\text{near}$ is  attempted to be rewired [line \ref{alg1:rew_check}, Alg. \ref{alg:1}], it may happen that $v_\text{near}.t$ changes so that we need to re-check \eqref{eq:rob_check} for $v_\text{near}$ as well as for all leaf nodes of $v_\text{near}$. For a time stamp $t\in\mathbb{N}_{\ge 0}$, the function $\textsc{CheckConstrRew}(v_\text{near},t)$ performs this step. 
Particularly, let $v_\text{near},v_1,\hdots,v_L$ be the sequence of nodes that defines the tree starting from node $v_\text{near}$ and ending in the leaf node $v_L$. Let $t$ be the new time stamp of $v_\text{near}$. In order to check \eqref{eq:rob_check} for these nodes, we change the time stamps of these nodes to $v_\text{near}.t=t$ and $v_l.t=t+l$ for $l\in\{1,\hdots,L\}$. We then check if $\textsc{CheckConstrRob}(v_\text{near})$ and $\textsc{CheckConstrRob}(v_l)$ hold for all $l\in\{1,\hdots,L\}$.

\textsc{TimeStampRew}: In the case that a node $v_\text{near}\in\mathcal{V}$ is rewired, we need to update the time stamps of $v_\text{near}$ and all its leaf nodes that we denote by $v_1,\hdots,v_L$. For a node $v_\text{near}$ and a time stamp $t\in\mathbb{N}_{\ge 0}$, the function $\textsc{TimeStampRew}(\mathcal{V},v_\text{near},t)$ changes the time stamps of these nodes to $v_\text{near}=t$ and $v_l.t=t+l$ for $l\in\{1,\hdots,L\}$ and outputs the modified set of nodes $\mathcal{V}$.

\begin{algorithm}
    \caption{Robust RRT$^*$: Tree Expansion}
    \begin{algorithmic}[1]
    \Procedure{$(\mathcal{V},\mathcal{V}_G,\mathcal{E})$=RRT$^*$($\mathcal{V}_0,\mathcal{V}_{G,0},\mathcal{E}_0,N$)}{}  
    \State $\mathcal{V} \gets \mathcal{V}_0$, $\mathcal{V}_G \gets \mathcal{V}_{G,0}$, $\mathcal{E} \gets \mathcal{E}_0$\label{alg1:initV}
        \For{$i = 1, \hdots, N$}  \label{alg1:forIter}                  
            \State $s_\text{rand} \gets \textsc{Sample}(i)$\label{alg1:sample}
            \State $v_\text{nearest} \gets \textsc{Nearest}(\mathcal{V},\mathcal{E},s_\text{rand} )$ \label{alg1:nearest}
            \State $(v_\text{new},u_\text{new}) \gets \textsc{Steer}(v_\text{nearest},s_\text{rand},\eta)$ \label{alg1:steer}
            \If{$\textsc{CheckConstrRob}(v_\text{new})$ } \label{alg1:CheckConstrRob}
                \State $s_\text{new}=(v_\text{new}.\mu,v_\text{new}.\Sigma)$
                \State $\mathcal{V}_\text{near} \gets \textsc{Near}(\mathcal{V},\mathcal{E},s_\text{new},\eta)$ \label{alg1:Vnear}
                \State $v_\text{min} \gets v_\text{nearest}$
                \State $c_\text{min} \gets \textsc{Cost}(v_\text{nearest})+c(v_\text{new}.\mu)$
                \For{$v_\text{near}\in V_\text{near}$}   
                \State $(v_\text{new}',u_\text{new}') \gets \textsc{Steer}(v_\text{near},s_\text{rand},\eta)$\label{alg1:steer_}
                    \If{$ \textsc{Cost}(v_\text{near})+c(v_\text{new}'.\hat{\mu})<c_\text{min}$}
                    \State $v_\text{min} \gets v_\text{near}$
                    \State $c_\text{min} \gets \textsc{Cost}(v_\text{near})+c(v_\text{new}'.\hat{\mu})$
                    \State $v_\text{new} \gets v_\text{new}'$
                    \State $u_\text{new} \gets u_\text{new}'$\label{alg1:end1}
                    \EndIf
                \EndFor
                \State $\mathcal{V} \gets \mathcal{V} \cup \{v_\text{new}\}$ \label{alg1:new_node}
                \State $\mathcal{E} \gets \mathcal{E} \cup \{(v_\text{min},v_\text{new}),u_\text{min}\}$\label{alg1:new_edge}
                \If{$\textsc{GoalReached}(v_\text{new})$}\label{alg1:GoalReached}
                \State ${\mathcal{V}_G}\gets {\mathcal{V}_G}\cup \{v_\text{new}\}$\label{alg1:updateGoalSet}
                \EndIf
                \For{$v_\text{near}\in V_\text{near}\setminus \{v_\text{new}\}$} \label{alg1:rew_start}
                \State $s_\text{near} \gets (v_\text{near}.\mu,v_\text{near}.\Sigma)$
                \State $(v_\text{new}',u_\text{new}') \gets \textsc{Steer}(v_\text{new},s_\text{near},\eta)$\label{alg1:steer__}
                    \If{$v_\text{near}.\mu=v_\text{new}'.\mu \wedge v_\text{near}.\Sigma=v_\text{new}'.\Sigma $\WRP{$\wedge \textsc{Cost}(v_\text{new})+c(v_\text{new}'.\hat{\mu})<\textsc{Cost}(v_\text{near})$}\WRP{$\wedge\textsc{CheckConstrRew}(v_\text{near},v_\text{new}'.t)$}}\label{alg1:rew_check}
                    \State $\mathcal{V} \gets \textsc{TimeStampRew}(\mathcal{V},v_\text{near},v_\text{new}'.t)$
                    \State $e_\text{parent} \gets \textsc{Parent}(v_\text{near})$
                    \State $\mathcal{V} \gets (\mathcal{V} \setminus \{v_\text{near}\}) \cup \{v_\text{new}'\}$
                    \State $\mathcal{V}_G \gets \mathcal{V}_G \setminus \{v_\text{near}\}$
                    \If{$\textsc{GoalReached}(v_\text{new}')$}
                    \State ${\mathcal{V}_G}\gets {\mathcal{V}_G}\cup \{v_\text{new}'\}$
                    \EndIf
                    \State $\mathcal{E} \gets (\mathcal{E} \setminus \{e_\text{parent}\}) \cup \{(v_\text{new},v_\text{new}'),u_\text{new}'\}$ \label{alg1:rew_end}
                    \EndIf
                \EndFor
            \EndIf
        \EndFor
    \EndProcedure
    \end{algorithmic}
    \label{alg:1}
\end{algorithm}


\subsection{Real-time Execution and Replanning}
\label{sec:online}
Having constructed a tree by means of Algorithm \ref{alg:1}, we can now find a control sequence $U_H$ from $\mathcal{V}_G\subseteq \mathcal{V}$ as described in Section \ref{sec:tree}. This sequence will satisfy the constraints of the optimization problem \eqref{eq:Prob2} as will be formally shown in Section \ref{sec:guarantees}. Towards solving the optimization problem \eqref{eq:Prob1}, we propose an online execution and replanning scheme in Algorithm \ref{alg:2}. A sequence $U_H$ is initially calculated in lines \ref{eq:alg2_tree}-\ref{eq:alg2_seq} and executed in lines \ref{eq:alg2_while}-\ref{eq:alg2_while_end}. Recall that the proposed R-RRT$^*$ in Algorithm \ref{alg:1} does not make use of $Y_t$ and instead introduces robustness margins $\epsilon$. There is hence a fundamental trade-off between the size of $\epsilon$ and the number of times $X(t|t)\not\in \mathcal{B}_{\epsilon(A,t)}(\hat{X}(t|t))$. In these cases, we trigger replanning in lines \eqref{eq:alg2_re}-\eqref{eq:alg2_re_}.
\begin{algorithm}
    \caption{Robust RRT$^*$: Real-time Execution}
    \begin{algorithmic}[1]
    \State $t \gets 0$
    \State $(\mathcal{V},\mathcal{V}_G,\mathcal{E}) \gets$ RRT$^*(v_0,\emptyset,\emptyset,N)$\label{eq:alg2_tree}
    \State Find control sequence $U_H$ from $\mathcal{V}_G\subseteq\mathcal{V}$\label{eq:alg2_seq}
    \While{$R(\|X(t|t)-O_1\|- \kappa)> \gamma$)}\label{eq:alg2_while}
        \State Collect measurement $y(t)$
        \State Apply $u(t)$
        \If{$X(t|t)\not\in \mathcal{B}_{\epsilon(A,t)}(\hat{X}(t|t))$}\label{eq:alg2_re}
            \State $(\mathcal{V},\mathcal{V}_G,\mathcal{E}) \gets$ RRT$^*(v_{t+1},\emptyset,\emptyset,N)$
            \State Find control sequence $U_H$ from $\mathcal{V}_G\subseteq\mathcal{V}$\label{eq:alg2_re_}
        \EndIf
        \State $t \gets t+1 $\label{eq:alg2_while_end}
    \EndWhile
    \end{algorithmic}
    \label{alg:2}
\end{algorithm}

\section{Theoretical Guarantees of R-RRT$^*$}
\label{sec:guarantees}
Let us first show soundness of our proposed method with respect to the constraints \eqref{constr2_}-\eqref{constr1b_}.
\begin{theorem}[Constraint Satisfaction of \eqref{eq:Prob2}]\label{thm:1}
Let the tree $(\mathcal{V},\mathcal{V}_G,\mathcal{E})=\text{RRT}^*(v_0,\emptyset,\emptyset,N)$ be obtained from Algorithm \ref{alg:1} for some $N\in\mathbb{N}_{\ge 0}$. Let $v_0,\hdots,v_H$ be a path in $(\mathcal{V},\mathcal{E})$ with $v_H\in\mathcal{V}_G$ and let $U_H$ be the associated control sequence, i.e., $(v_t,v_{t+1},u(t))\in\mathcal{E}$ for all $t\in\{0,\hdots,H-1\}$. Then it holds that the constraints \eqref{constr2_}-\eqref{constr1b_} are satisfied.

\begin{proof}
Note first that every node $v_t$ is such that
\begin{align}\label{eq:cons_proof}
\sup_{{X} \in\mathcal{B}_{\epsilon(A,v_t.t)}(\hat{X}_t)}R(-d_j({X},O_2,\hdots,O_M))\leq\gamma_j, \forall j\in\ccalJ
\end{align}
where $\hat{X}_t\sim\mathcal{N}(v_t.\mu,v_t.\Sigma)$. This follows because every node $v_\text{new}$ that is added to the tree in line \ref{alg1:new_node} of Algorithm \ref{alg:1} is checked for \eqref{eq:cons_proof} via $\textsc{CheckConstrRob}$ in line \ref{alg1:CheckConstrRob}. Also note that the function  $\textsc{CheckConstrRew}$ in line \ref{alg1:rew_check} ensures that after rewiring each node $v.t$ still ensures \eqref{eq:cons_proof}. Similarly, the node $v_H$ satisfies the goal constraint
\begin{align}\label{eq:cons_2_proof}
\sup_{{X} \in\mathcal{B}_{\epsilon(A,v_H.t)}(\hat{X}_H)}R(\|\hat{X}-O_1\|-\kappa)\leq\gamma
\end{align}
due to the function $\textsc{GoalReached}$ in line \ref{alg1:GoalReached}. Since the \textsc{Steer} function,  called in lines \ref{alg1:steer}, \ref{alg1:steer_}, and \ref{alg1:steer__} of Alg. \ref{alg:1}, ensures that the constraints \eqref{eq:steer1}, \eqref{eq:steer2}, and \eqref{eq:steer3} hold, it consequently follows that \eqref{constr2_}-\eqref{constr1b_} are satisfied. 
\end{proof}
\end{theorem}

Let us next show that Algorithm \ref{alg:1} has the property that the cost of each node decreases as we keep growing the tree.
\begin{theorem}[Non-increasing Cost Function \eqref{obj2_}]
Let the tree $(\mathcal{V},\mathcal{V}_G,\mathcal{E})=RRT^*(v_0,\emptyset,\emptyset,N)$ be obtained from Algorithm \ref{alg:1} for some $N\in\mathbb{N}_{\ge 0}$. If we  extend this tree by calling $(\mathcal{V}',\mathcal{V}_G',\mathcal{E}')=\text{RRT}^*(\mathcal{V},\mathcal{V}_G,\mathcal{E},1)$, then it holds that $\textsc{Cost}(v')\le \textsc{Cost}(v)$ for each $v\in \mathcal{V}$ and $v'\in \mathcal{V}'$ with $v.\mu=v'.\mu$ and $v.\Sigma=v'.\Sigma$.

\begin{proof}
The proof follows by construction of the rewiring in Algorithm \ref{alg:1} as nodes are rewired only if their cost decreases after rewiring as per line [\ref{alg1:rew_check}, Alg. \ref{alg:1}].
\end{proof}
\end{theorem}

\begin{remark}[Optimality]
Note that proving global asymptotic  optimality, as in case of  RRT$^*$ \cite{karaman2010optimal}, is an open problem as our incremental tree construction depends on time. 
\end{remark}

Let us next analyze in what way our solution to the optimization problem \eqref{eq:Prob2} relates to solving the optimization problem \eqref{eq:Prob1}. Let us first state a straightforward corollary.

\begin{cor}[Constraint Satisfaction of \eqref{eq:Prob1}]\label{cor.1}
Let the tree $(\mathcal{V},\mathcal{V}_G,\mathcal{E})=\text{RRT}^*(v_0,\emptyset,\emptyset,N)$ be obtained from Algorithm \ref{alg:1} for some $N\in\mathbb{N}_{\ge 0}$. Let $v_0,\hdots,v_H$ be a path in $(\mathcal{V},\mathcal{E})$ with $v_H\in\mathcal{V}_G$ and let $U_H$ be the associated control sequence, i.e., $(v_t,v_{t+1},u(t))\in\mathcal{E}$ for all $t\in\{0,\hdots,H-1\}$. Let $\hat{X}(t|t)\sim\mathcal{N}({\mu}(t|0),{\Sigma}(t|t))$ be a random variable and let the realized disturbance $y(t)$ be such that
    \begin{align*}
        X(t|t)\in \mathcal{B}_{\epsilon(A,t)}(\hat{X}(t|t))
    \end{align*}
for all $t\in\{1,\hdots,H\}$. Then it holds that \eqref{constr2b} and \eqref{constr1b} hold.

\begin{proof}
Follows by the proof of Theorem \ref{thm:1}, and in particular due to the satisfaction of \eqref{eq:cons_proof} and \eqref{eq:cons_2_proof}.
\end{proof}
\end{cor}

Note that larger $\epsilon$ will increase the probability of satisfying the constraints of \eqref{eq:Prob1} as it is more likely that $X(t|t)\in \mathcal{B}_{\epsilon(A,t)}(\hat{X}(t|t))$. Let us next quantify the probability $\delta\in[0,1]$ such that $X(t|t)\in \mathcal{B}_{\epsilon(A,t)}(\hat{X}(t|t))$ for all $t\in \{1,\hdots,H\}$. Recall that $X(t|t)\in\mathcal{N}(\mu(t|t),\Sigma(t|t))$. From \eqref{eq:kalman_eq}, note that $\mu(t|t)$ is a random variable with a mean and a covariance when $y(t)$ is treated instead as a random variable $Y(t)$. Denote this random variable by $M(t|t)$ and note that
\begin{align*}
    M(t+1|t+1)&=F_\mu(M(t|t),u(t),Y(t+1))\\
    &=F_\mu(M(t|t),u(t),CX(t+1)+V(t+1))
\end{align*}
We have that $M(t|t)$ is linear in $X(t)$ and $V(t)$, which both follow a Gaussian distribution, so that $M(t|t)$ again follows a Gaussian distribution. We can then calculate
\begin{align*}
   \delta&:= P(X(t|t)\in \mathcal{B}_{\epsilon(A,t)}(\hat{X}(t|t)),  t\in\{1,\hdots,H\})\\
   &=P(\|M(t|t)-{\mu}(t|0)\|^2\le \epsilon(A,t) ,  t\in\{1,\hdots,H\})
\end{align*}
as the probability that $X(t|t)\in \mathcal{B}_{\epsilon(A,t)}(\hat{X}(t|t))$.
\begin{cor}
Let the tree $(\mathcal{V},\mathcal{V}_G,\mathcal{E})=RRT^*(v_0,\emptyset,\emptyset,N)$ be obtained from Algorithm \ref{alg:1} for some $N\in\mathbb{N}_{\ge 0}$. Let $v_0,\hdots,v_H$ be a path in $(\mathcal{V},\mathcal{E})$ with $v_H\in\mathcal{V}_G$ and let $U_H$ be the associated control sequence, i.e., $(v_t,v_{t+1},u(t))\in\mathcal{E}$ for all $t\in\{0,\hdots,H-1\}$. Then with a probability $\delta$ the constraints \eqref{constr2b} and \eqref{constr1b} hold.

\begin{proof}
Follows by Theorem \ref{thm:1} and Corollary \ref{cor.1}.
\end{proof}
\end{cor}
\begin{rem}
Our proposed method hence allows to make statements such as ``with a probability of  $\delta$, the constraints \eqref{constr2b} and \eqref{constr1b} will hold''. The probability $\delta$ naturally increases with the size of $\epsilon$, which increases conservatism.
\end{rem}

We now suggest how to potentially select $\epsilon$. Given two nodes $v,v'\in \mathcal{V}$, recall that the node $v$ encodes the set of distributions $X\in \mathcal{B}_{\epsilon(A,v.t)}(v.{\mu},v.{\Sigma})$. If there exists an edge between $v$ and $v'$, i.e., $(v,v',u)\in \mathcal{E}$, a desirable property is that there exists a dynamically feasible transition from each distribution  $X\in\mathcal{B}_{\epsilon(A,v.t)}(v.{\mu},v.{\Sigma})$ such that $AE(X)+Bu+\mu_W\in\mathcal{B}_{\epsilon(A,v'.t)}(v'.{\mu},v'.{\Sigma})$  to not trigger replanning too frequently. Note that $E(X)$ is the expected value of $X$. Hence, for each $X\in \mathcal{B}_{\epsilon(A,v.t)}(v.{\mu},v.{\Sigma})$,
\begin{align}\label{eq:dyn_rob}
   \|AE(X) +Bu + \mu_W - v'.{\mu}\|^2 \le \epsilon(A,v'.t)
\end{align}
has to hold. We show conditions  under which  \eqref{eq:dyn_rob} holds.
\begin{theorem}
Let $\epsilon(A,t)=\vertiii{A}^t\zeta$ for some $\zeta>0$. Then for a transition $(v,v',u)\in \mathcal{E}$, there exists a dynamically feasible transition from each distribution in $\mathcal{B}_{\epsilon(A,v.t)}(v.{\mu},v.{\Sigma})$ into $\mathcal{B}_{\epsilon(A,v'.t)}(v'.{\mu},v'.{\Sigma})$, i.e., \eqref{eq:dyn_rob} holds.

\begin{proof}
Note that the control input $u$ is such that the transition $(v.{\mu},v.{\Sigma})$ into $(v'.\mu,v'.\Sigma)$ is dynamically feasible, i.e., such that $v'.\mu=Av.{\mu}+Bu+\mu_W$. For any $\mathcal{N}(\mu,\Sigma)\in \mathcal{B}_{\epsilon(A,v.t)}(v.{\mu},v.{\Sigma})$, let $\mu':=A\mu+Bu+\mu_W$. We now have that $\|v'.\mu-\mu'\|=\|A(v.{\mu}-\mu)\|\le \vertiii{A}\|v.{\mu}-\mu\|\le \vertiii{A}\epsilon(A,v.t)=\vertiii{A}\vertiii{A}^{v.t}\zeta=\vertiii{A}^{v'.t}\zeta= \epsilon(A,v'.t)$. 
\end{proof}
\end{theorem}



\section{Simulation Studies}
\label{sec:experiments}


We next demonstrate our proposed robust RRT$^*$ algorithm. In Section \ref{sec:simDyn}, we define the stochastic system dynamics as per \eqref{eq:rdynamics} that we will use throughout this section along with the environment in which the system operates. In Section \ref{sec:tunnels}, we illustrate the effect of the robustness parameter $\epsilon$ on the path design. Specifically, we illustrate the aforementioned trade-off between robustness and conservatism, and we show that as $\epsilon$ decreases, the re-planning frequency increases (see Alg. \ref{alg:2}). In this way, we illustrate how $\epsilon$ can be used as a design parameter to do planning in between very optimistic and very conservative planning.

\subsection{System Dynamics \& Environment}\label{sec:simDyn}
\begin{figure*}
    \centering
    \begin{subfigure}[b]{0.32\textwidth}
        \centering
        \includegraphics[width=\textwidth]{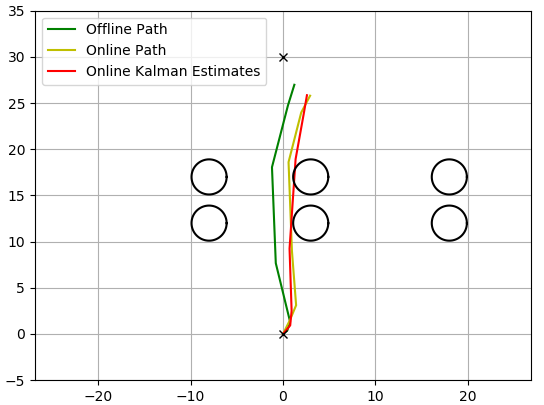}
        \caption{$\eps=0$ (no replanning)} 
        \label{fig:1}
    \end{subfigure}
    \begin{subfigure}[b]{0.32\textwidth}
        \centering
        \includegraphics[width=\textwidth]{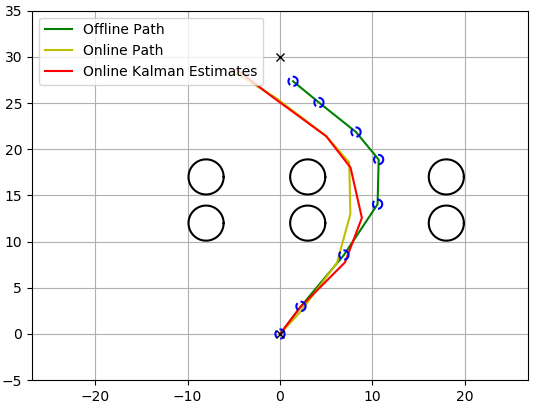}
        \caption{$\eps=0.5$ (no replanning)} 
        \label{fig:2}
    \end{subfigure}
    \begin{subfigure}[b]{0.32\textwidth}  
        \centering 
        \includegraphics[width=\textwidth]{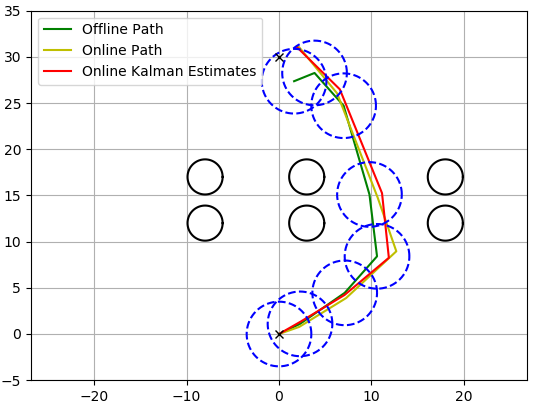}
        \caption{$\eps=3.5$ (no replanning)}  
        \label{fig:3}
    \end{subfigure}\\
    \begin{subfigure}[b]{0.32\textwidth}   
        \centering 
        \includegraphics[width=\textwidth]{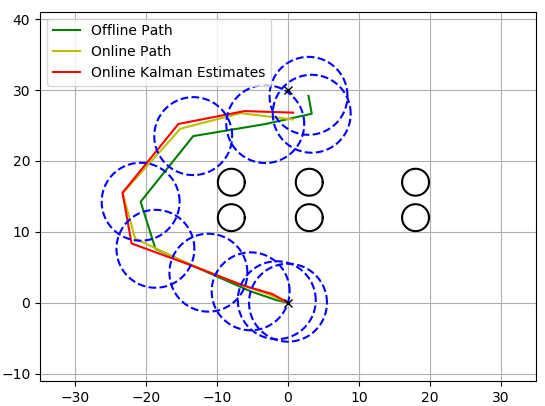}
        \caption{$\eps=5.5$ (no replanning)}
        \label{fig:4}
    \end{subfigure}
        \begin{subfigure}[b]{0.32\textwidth}
        \centering
        \includegraphics[width=\textwidth]{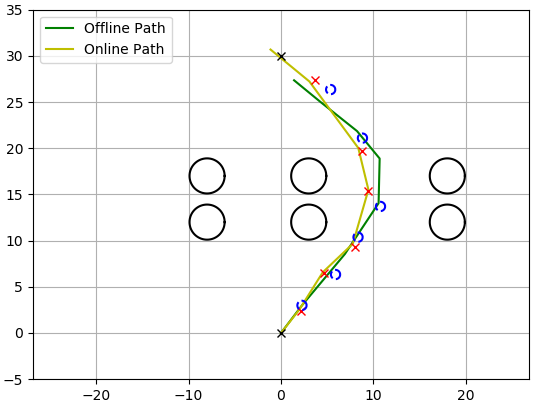}
        \caption{$\eps=0$ (with replanning)} 
        \label{fig:5}
    \end{subfigure}
    \begin{subfigure}[b]{0.32\textwidth}
        \centering
        \includegraphics[width=\textwidth]{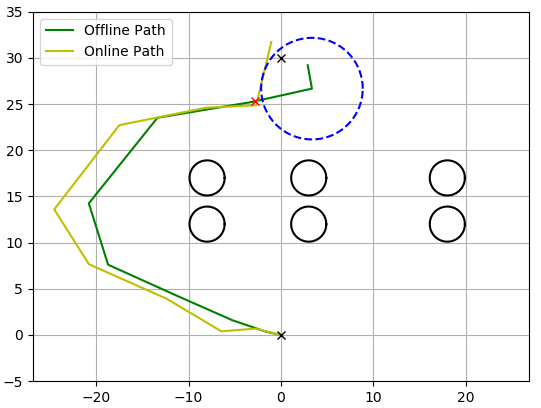}
        \caption{$\eps=5.5$ (with replanning)} 
        \label{fig:6}
    \end{subfigure}
    \caption{Simulation results with constant $\epsilon$. (a)-(d) show the open-loop policy, while (e)-(f) show the replanning according to Algorithm \ref{alg:2}. The blue dashed circles indicate the epsilon balls for some time steps around the offline paths. The green line is the nominal mean $\mu(t|0)$, i.e., the path as obtained from Algorithm \ref{alg:1}, the red line is the realized Kalman estimate $\mu(t|t)$ observed during execution of Algorithm \ref{alg:2}, and the yellow line is the realized path $x(t)$ of the robot.}
    \label{fig:7}
\end{figure*}

Consider the stochastic system as in \eqref{eq:rdynamics} defined by
\begin{align*}
    A:=\mathcal{I}_2\otimes \begin{bmatrix}
    1 & 0.5  \\
    0 & 1  
    \end{bmatrix} \;  B :=\mathcal{I}_2\otimes \begin{bmatrix}
    0.125  \\
    0.5
    \end{bmatrix} \;
    C := \mathcal{I}_2\otimes\begin{bmatrix}
    1 & 0 
    \end{bmatrix} \nonumber
\end{align*}
where $\mathcal{I}_2\in\mathbb{R}^{2\times 2}$ is the identity matrix and $\otimes$ is the Kronecker product. The state $x:=(x_1,v_1,x_2,v_2)$ consists of position and velocity in the first and second coordinate. This system describes discretized two-dimensional double integrator dynamics with a sampling time of $0.5$ s, e.g., a service robot navigating through an obstacle cluttered environment. The process and measurement noise are such that $W(t)\sim\mathcal{N}\Big(\mathcal{O}_4,\mathcal{I}_2\otimes \begin{bmatrix}
    0.5 & 0.1  \\
    0.1 & 0.5 
    \end{bmatrix}\Big)$ and $V(t)\sim\mathcal{N}\big(\mathcal{O}_2,\mathcal{I}_2\otimes 0.4
    \big)$ where $\mathcal{O}_i$ denotes an $i$-dimensional vector containing zeros.  
The robot operates in the environment shown in Figure \ref{fig:1}, where the initial and goal location of the robot are $X(0):=(0,0,0,0)$ and $O_1=(0,0,30,0)$, respectively. Observe also that there are three corridors for the robot to traverse through as defined by the obstacles $O_2,\hdots,O_7$ that are indicated by the black circles. Importantly, note that these corridors have different width. We consider $J:=6$ risk constraints and let each function be $d_j(X,O_{j+1}):=\|X-O_{j+1}\|$ for  $j\in\{1,\hdots,6\}$. We select $\gamma:=\gamma_j:=-0.5$ and $\kappa:=0.5$ and use the conditional value-at-risk  $R(\cdot):=CVaR_\beta(\cdot)$ at risk level $\beta:=0.95$. To check \eqref{eq:rob_check} and \eqref{eq:rob_check_} for a node $v\in\mathcal{V}$, we sample $10$ Gaussian distributions $X\in \mathcal{B}_{\epsilon(A,v.t)}(\hat{X})$ where we recall that $\hat{X}\sim \mathcal{N}(v.\mu,v.\Sigma)$. Then, we check for each $X$ if $R(-d_j(X,O_2,\hdots,O_M))\leq\gamma_j, \forall j\in\ccalJ$ and $R(\|X-O_1\|-\kappa)\leq\gamma$ hold, respectively, using $500$ samples from $X$ and a sample average approximation \cite{rockafellar2000optimization}. For  $\epsilon$  of $0$, $0.5$, $3.5$, and $5.5$ (see next section), we observed average run times, i.e., until a satisfying path was found  by Algorithm \ref{alg:1}, of $114$, $127$, $309$, and $288$ s. We observed that checking \eqref{eq:rob_check} and \eqref{eq:rob_check_} in Algorithm \ref{alg:1} is computationally expensive, and remark that we plan to derive efficient reformulations in the future.  



\subsection{Effect of Robustness on Path Design}\label{sec:tunnels}

\begin{figure*}
    \centering
    \begin{subfigure}[b]{0.32\textwidth}
        \centering
        \includegraphics[width=\textwidth]{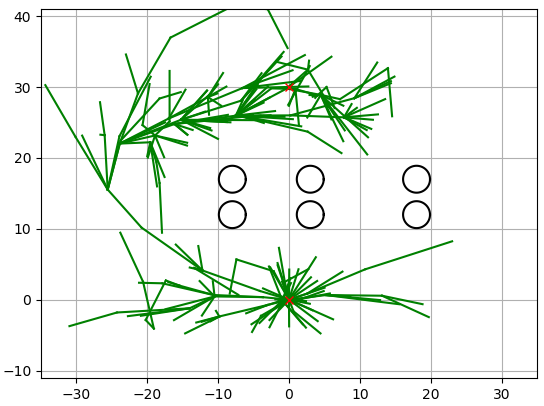}
        \caption{$\eps=5.5$} 
        \label{fig:t2}
    \end{subfigure}
    \begin{subfigure}[b]{0.32\textwidth}
        \centering
        \includegraphics[width=\linewidth]{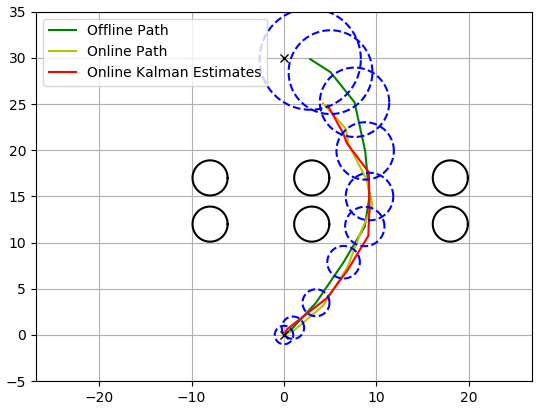}
        \caption{$\epsilon(A,t)=\vertiii{A}^t \zeta$ and $\zeta=1$} 
        \label{fig:s1}
    \end{subfigure}
    \begin{subfigure}[b]{0.32\textwidth}
        \centering
        \includegraphics[width=\linewidth]{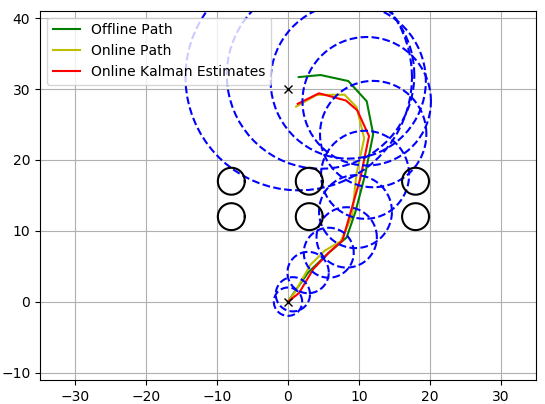}
        \caption{$\epsilon(A,t)=\vertiii{A}^t \zeta$ and $\zeta=2$} 
        \label{fig:s2}
    \end{subfigure}
    \caption{(a) shows a grown tree of states for a robustness $\epsilon=5.5$, while (b)-(c) show results for time-varying $\epsilon(A,t)$.}
    \label{fig:s3}
\end{figure*}

For our proposed R-RRT$^*$, we select $N:=700$ and first consider constant $\epsilon$ of different sizes. In Figs. \ref{fig:1}-\ref{fig:4} we show the result when no replanning is considered, i.e., Algorithm~\ref{alg:2} is run without lines \ref{eq:alg2_re}-\ref{eq:alg2_re_} so that the open-loop policy is executed for choices of $\epsilon=0,0.5,3.5,5.5$. It can be observed that increasing $\epsilon$ naturally results in selecting the path that allows safer distance to the obstacles at the expense of having a larger cost over the planned path. Note also that for $\epsilon=0$ the robot collides with one of the obstacles. In Figs. \ref{fig:5} and \ref{fig:6}, Algorithm \ref{alg:2} is run with replanning as indicated by the red crosses. It can be observed that smaller $\epsilon$ result in more frequent replanning. In Fig. \ref{fig:t2}, we show the grown trees in green for $\epsilon=5.5$. In Figs. \ref{fig:s1}-\ref{fig:s2} we  used time-varying epsilon, i.e, $\epsilon(A,t)$ to account for growing estimation uncertainty as time increases.  

Finally, let us remark that a comparison with a version of a stochastic RRT$^*$ that does not incorporate measurements, such as for instance in \cite{luders2010chance}, was not possible as the planning problem did initially not find a feasible solution after $N:=700$ iterations of sampling new nodes. The reason here is that the unconditional covariance matrix $\Sigma(t)$, which is used for planning, grows unbounded.

\section{Conclusions and Future Work}
\label{conclusion}
We considered the  robust motion planning problem in the presence of state  uncertainty. In particular, we proposed a novel sampling-based approach that introduces robustness margins into the offline planning to account for uncertainty in the state estimates based on a Kalman filter. We complement the robust offline planning with an online replanning scheme and show an inherent trade-off in the size of the robustness margin and the frequency of replanning. Future work includes integration of perception and feedback control.

\bibliographystyle{IEEEtran}
\bibliography{YK_bib.bib}

\appendix
\section{Risk Measures}
\label{app:risk}
We next present some desireable properties that a risk measure may have. Let therefore $Z,Z'\in \mathfrak{F}(\Omega,\mathbb{R})$ be random variables. A risk measure is  \emph{coherent} if the following four properties are satisfied.\\
\emph{1. Monotonicity:} If $Z(\omega) \leq Z'(\omega)$ for all $\omega\in\Omega$, it holds that $R(Z) \le R(Z')$.\\
\emph{2. Translation Invariance:} Let $c\in\mathbb{R}$. It holds that $R(Z + c) = R(Z) + c$.\\
\emph{3. Positive Homogeneity:} Let $c\in\mathbb{R}_{\ge 0}$. It holds that $R(c Z) =  R(Z)$.\\
\emph{4. Subadditivity:} It holds that $R(Z + Z') \leq R(Z) + R(Z')$.

If the risk measure additionally satisfies the following two properties, then it is called a distortion risk measure. 
\emph{5. Comonotone Additivity:} If $(Z(\omega) - Z(\omega'))(Z'(\omega) - Z'(\omega')) \ge 0$ for all $\omega, \omega' \in \Omega$ (namely, $Z$ and $Z'$ are commotone), it holds that $R(Z + Z') = R(Z) + R(Z')$.\\
\emph{6. Law Invariance:} If $Z$ and $Z'$ are identically distributed, then $R(Z) = R(Z')$.

Common examples of popular risk measures are the expected value $\text{E}(Z)$ (risk neutral) and the worst-case $\text{ess} \sup_{\omega\in\Omega} Z(\omega)$ as well as:
\begin{itemize}
	\item Mean-Variance: $\text{E}(Z) + \lambda \text{Var}(Z)$ where  $c> 0$.
	\item Value at Risk (VaR) at level $\beta \in (0,1)$: $VaR_\beta(Z):=\inf \{ \alpha \in \mathbb{R} |  F_Z(\alpha) \ge \beta \}$.
	\item Conditional Value at Risk (CVaR) at level $\beta \in (0,1)$: $CVaR_\beta(Z):=E(Z|Z>VaR_\beta(Z))$.
\end{itemize}

Many risk measures are not coherent and can lead to a misjudgement of risk, e.g., the mean-variance is not monotone and the  value at risk (which is closely related to chance constraints as often used in optimization) does not satisfy the subadditivity property. 

\section{State Estimation}
\label{app:state_est}
The random variable $X(t)$ of the stochastic control system in \eqref{eq:rdynamics} is defined by the unconditional mean $\mu(t):=E[X(t)]$ and the unconditional covariance matrix $\Sigma(t):=E[(X(t)-\mu(t))(X(t)-\mu(t))^T]$  can recursively be calculated as
\begin{align*}
    \mu(t+1)&=A\mu(t)+B u(t)+\mu_W,\\
    \Sigma(t+1)&=A\Sigma(t)A^T+ \Sigma_W.
\end{align*}

The random variable $X(t|s)$ of the stochastic control system in \eqref{eq:rdynamics} is defined by the conditional mean $\mu(t|s):=E[X(t)|Y_{s}]=E[X(t|s)]$ and the conditional covariance matrix $\Sigma(t|s):=E[(X(t|s)-\mu(t|s))(X(t|s)-\mu(t|s))^T]$. These
can again recursively be calculated by means of the Kalman filter using the prediction equations
\begin{align*}
    \mu(t+1|t)&:=A\mu(t|t)+Bu(t)+\mu_W\\
    \Sigma(t+1|t)&:=A\Sigma(t|t)A^T+\Sigma_W
\end{align*}
and the update equations
\begin{align*}
    \mu(t|t)&=\mu(t|t-1)+K(t)(y(t)-C\mu(t|t-1)-\mu_V)\\
    \Sigma(t|t)&=\Sigma(t|t-1)-K(t)C\Sigma(t|t-1)
\end{align*}
and the optimal Kalman gain 
\begin{align*}
    K(t):=\Sigma(t|t-1)C^T(C\Sigma(t|t-1)C^T+\Sigma_V)^{-1}.
\end{align*}
The prediction and update equations together with the Kalman gain $K(t)$ define the functions $F_\mu$ and $F_\Sigma$.

\end{document}